\documentclass[runningheads]{llncs}
\usepackage[T1]{fontenc}
\usepackage{graphicx}
\usepackage{booktabs}
\usepackage[misc]{ifsym}

\usepackage{bm}
\usepackage[export]{adjustbox}
\usepackage{amsmath,amssymb}
\usepackage{amsfonts}
\usepackage{siunitx}
\usepackage{algorithm}
\usepackage{subfigure}
\usepackage{wrapfig}
\usepackage{xcolor}
\usepackage{algpseudocode}
\algrenewcommand\algorithmicrequire{\textbf{Input:}}
\algrenewcommand\algorithmicensure{\textbf{Output:}}
\algnewcommand{\LineComment}[1]{\State \(\triangleright\) #1}
\usepackage{hyperref}
\hypersetup{colorlinks,allcolors=black}

\begin{document}

\title{Retrieval-Augmented Mining of Temporal Logic Specifications from Data}


\author{Gaia Saveri\inst{1, 2} \and Luca Bortolussi\inst{1}}

\authorrunning{Saveri G. et Bortolussi L.}

\institute{University of Trieste, Italy \and University of Pisa, Italy}


\maketitle              

\begin{abstract} 
The integration of cyber-physical systems (CPS) into everyday life raises the critical necessity of ensuring their safety and reliability. An important step in this direction is requirement mining, i.e. inferring formally specified system properties from observed behaviors, in order to discover knowledge about the system. Signal Temporal Logic (STL) offers a concise yet expressive language for specifying requirements, particularly suited for CPS, where behaviors are typically represented as time series data. This work addresses the task of learning STL requirements from observed behaviors in a data-driven manner, focusing on binary classification, i.e. on inferring properties of the system which are able to discriminate between regular and anomalous behaviour, and that can be used both as classifiers and as monitors of the compliance of the CPS to desirable specifications. We present a novel framework that combines Bayesian Optimization (BO) and Information Retrieval (IR) techniques to simultaneously learn both the structure and the parameters of STL formulae, without restrictions on the STL grammar. Specifically, we propose a framework that leverages a dense vector database containing semantic-preserving continuous representations of millions of formulae, queried for facilitating the mining of requirements inside a BO loop. We demonstrate the effectiveness of our approach in several signal classification applications, showing its ability to extract interpretable insights from system executions and advance the state-of-the-art in requirement mining for CPS.

\end{abstract}


\section{Introduction}\label{sec:intro}
The adoption of cyber-physical systems (CPS) is pervasive nowadays: wearable devices, internet of things applications and domotics systems are part of our daily lives. However, formal specifications describing their expected behaviour are not always (completely) available. This is extremely serious when CPS are deployed in safety-critical scenarios, such as autonomous vehicles or medical devices, where safety and reliability are key desiderata. Requirement (or specification) mining is the the task of inferring system properties from observations of its behaviour~\cite{stl-mining-survey}. In the context of CPS, such behaviours are often recorded as time series describing the dynamics of the system over time (e.g. readings from a sensor at regular time intervals), hence Signal Temporal Logic (STL) \cite{stl,temporal-logic} is a popular language for expressing requirements~\cite{stl-cps}. STL is indeed a formalism which allows to reason over real-valued trajectories in a concise yet rich way: for example in STL one can state properties like "the temperature of the room will reach $25$ degrees within the next $10$ minutes and will stay above $22$ degrees for the next hour". 
In this work, we tackle the objective of learning STL requirements from observed behaviours of a system, in a data-driven approach: differently from formal methods algorithms, our proposed solution does not require a complete model of the system, but only a fixed dataset of recorded trajectories. 
Specifically, we frame the STL requirement mining problem in the binary classification scenario, i.e. our goal is to learn a STL formula which is able to discriminate between regular and anomalous trajectories. In this context the mined STL specification can be used both as a classifier and as a monitor for the system at hand, and leveraged for maintenance and bug detection as well as a tool for system modeling and comprehension.  
\paragraph{Our contribution} consists in proposing a novel framework for STL specification mining, which involves combining Bayesian Optimization (BO) and  Information Retrieval (IR) techniques, to simultaneously learn the structure and the parameters of the specifications, without neither the need of a (generative) model of the system under analysis, nor of restrictions on the STL grammar to some specific fragment. Indeed we: (i) leverage the semantics of STL for mapping formulae in a continuous semantic-preserving latent space, using the method of \cite{stl-kernel}, see Section~\ref{sec:background}; (ii) build a dense vector database containing the embeddings of millions of formulae, carefully constructed in an incremental way, as detailed in Section~\ref{subsec:db}. Notably, this step is done once, and the same database is queried in all the application cases; (iii) frame the requirement mining task as an optimization problem in the semantic space of STL formulae, using queries to the database constructed at the previous step for inverting the embeddings, i.e. mapping continuous semantic vectors back to formulae (Sections \ref{sec:method}). 
We show the effectiveness of the proposed approach in several binary classification applications, proving that it is able to extract interpretable information from system executions, offering insights on the system under analysis. 

\section{Background}\label{sec:background}

\paragraph{Signal Temporal Logic} (STL) is a linear-time temporal logic which expresses properties on trajectories over dense time intervals \cite{stl}. We define as trajectories the functions $\xi: I\rightarrow D$, where $I\subseteq \mathbb{R}_{\geq 0}$ is the time domain and $D\subseteq \mathbb{R}^k, k\in \mathbb{N}$  is the state space. The syntax of STL is given by:
$$\varphi:=tt\mid\pi\mid\lnot\varphi\mid \varphi_1\land\varphi_2\mid\varphi_1\mathbf{U}_{[a, b]}\varphi_2$$
where $tt$ is the Boolean \emph{true} constant; $\pi$ is an \emph{atomic predicate}, i.e.\ a function over variables $\bm{x}\in \mathbb{R}^n$ of the form $f_{\pi}(\bm{x})\geq \num{0}$ (we refer to $n$ as the number of variables of a STL formula); $\lnot$ and $\land$ are the Boolean \emph{negation} and \emph{conjunction}, respectively (from which the \emph{disjunction} $\lor$ follows by De Morgan's law); $\mathbf{U}_{[a, b]}$, with $a, b \in \mathbb{Q}, a<b$, is the \emph{until} operator, from which the \emph{eventually} $\mathbf{F}_{[a, b]}$ and the \emph{always} $\mathbf{G}_{[a, b]}$ temporal operators can be deduced. We can intuitively interpret the temporal operators over $[a, b]$ as follows: a property is \textit{eventually} satisfied if it is satisfied at some point inside the temporal interval, while a property is \textit{globally} satisfied if it is true continuously in $[a, b]$; finally the \textit{until} operator captures the relationship between two conditions $\varphi, \psi$ in which the first condition $\varphi$ holds until, at some point in $[a, b]$, the second condition $\psi$ becomes true. We call $\mathcal{P}$ the set of well-formed STL formulae. STL is endowed with both a \emph{qualitative} (or Boolean) semantics, giving the classical notion of satisfaction of a property over a trajectory, i.e. $s(\varphi, \xi, t) = \num{1}$ if the trajectory $\xi$ at time $t$ satisfies the STL formula $\varphi$, and a  \emph{quantitative} semantics, denoted by $\rho(\varphi, \xi, t)$. The latter, also called \emph{robustness}, is a measure of how robust is the satisfaction of $\varphi$ w.r.t. perturbations of the signals. Robustness is recursively defined as:
\begin{small}
\begin{align*}
 & \rho(\pi,\xi,t) &=& f_\pi(\xi(t)) \qquad \text{for } \pi(\bm{x})=\big(f_\pi(\bm{x})\geq 0\big)\\
 & \rho(\lnot\varphi,\xi,t) &=& -\rho(\varphi,\xi,t)\\
  & \rho(\varphi_1\land\varphi_2,\xi,t) &=& \min\big(\rho(\varphi_1,\xi,t),  \rho(\varphi_2,\xi,t)\big)\\
 &\rho(\varphi_1\mathbf{U}_{[a, b]}\varphi_2,\xi,t) \hspace*{-0.5em}&=&  \max_{{t'\in[t+a,t+b]}}\big(\min\big(\rho(\varphi_2,\xi,t'),
 \min_{{t''\in[t,t']}}\rho(\varphi_1,\xi,t'')\big)\big) 
\end{align*}
\end{small}
Since they will be used later, we report also the definition of robustness of derived temporal operators: eventually $\rho(\mathbf{F}_{[a, b]}\varphi, \xi, t) = \max_{t'\in[t+a,t+b]} \rho(\varphi,\xi,t)$ and globally $\rho(\mathbf{G}_{[a, b]}\varphi, \xi, t) = \min_{t'\in[t+a,t+b]} \rho(\varphi,\xi,t)$. Robustness is compatible with satisfaction via the following \emph{soundness} property: if $\rho(\varphi, \xi, t) > 0$ then $s(\varphi, \xi, t) = \num{1}$ and if $\rho(\varphi, \xi, t) < \num{0}$ then $s(\varphi, \xi, t) = \num{0}$. When $\rho(\varphi, \xi, t) = \num{0}$ arbitrary small perturbations of the signal might lead to changes in satisfaction value. We omit $t$ from the previous notations when properties are evaluated at time $t=\num{0}$. 

In this context stochastic processes are probability spaces defined as triplets $\bm{X} = (\mathcal{T}, \mathcal{A}, \mu)$ of a trajectory space $\mathcal{T}$ and a probability measure $\mu$ on a $\sigma$-algebra $\mathcal{A}$ over $\mathcal{T}$. Given a STL formula $\varphi$ with predicates interpreted over state variables of $\bm X$ and a trajectory of the stochastic system $\xi(t)$, its robustness $\rho(\varphi, \xi)$ is a functional $R_{\varphi}: \mathcal{T}\rightarrow \mathbb{R}$, which defines the real-valued random variable $R_{\varphi} = R_{\varphi}(\bm{X})$,  following the distribution:
\begin{equation}
    \mathbb{P}(R_{\varphi}(\bm{X})\in [a, b]) = \mathbb{P}(\bm{X}\in \{\xi \in \mathcal{T} | \rho(\varphi, \xi)\in [a, b]\})
    \label{eq:stochrob}
\end{equation}
The expected value of the previous, i.e. $\mathbb{E}[R_{\varphi}|\bm{X}]$, intuitively gives a measure of how strongly a formula is satisfied by trajectories sampled from the process $\bm{X}$ (the higher the value the more robust the satisfaction) \cite{expected-rob}; in the remainder of this paper, this expectation will be approximated by Monte Carlo sampling.

\paragraph{Finding continuous representations} of STL formulae is performed with an ad-hoc kernel in \cite{stl-kernel}, by leveraging the quantitative semantics of STL. Indeed, robustness allows formulae to be considered as functionals mapping trajectories into real numbers, i.e.\ $\rho(\varphi,\cdot): \mathcal{T}\rightarrow \mathbb{R}$ such that $\xi\mapsto \rho(\varphi, \xi)$. Considering these as feature maps, and fixing a probability measure $\mu_0$ on the space of trajectories $\mathcal{T}$, a kernel function capturing similarity among STL formulae on mentioned feature representations can be defined as:
\begin{equation}
k(\varphi, \psi) = \langle \rho(\varphi, \cdot), \rho(\psi, \cdot) \rangle = \int_{\xi\in \mathcal{T}} \rho(\varphi, \xi) \rho(\psi, \xi) d\mu_0(\xi)
\label{eq:stl-kernel}
\end{equation}
opening the doors to the use of the scalar product in the Hilbert space $L^2$ as a kernel for $\mathcal{P}$; intuitively this results in a kernel having high positive value for formulae that behave similarly on high-probability trajectories (w.r.t. $\mu_0$), and viceversa low negative value for formulae that on those trajectories disagree. 
\begin{wrapfigure}{R}{0.5\textwidth}
\begin{minipage}{0.5\textwidth}
\vspace{-2.1cm}
\begin{algorithm}[H]
\caption{Sampling a trajectory over the interval $[a, b]$ according to $\mu_0$}
\label{alg:mu0}
\begin{algorithmic}
\Require $\Delta$, $a$, $b$, $m'$, $m''$, $\sigma'$, $\sigma''$, $q$
\Ensure $\xi$ 
\LineComment sample the starting point  
\State $\xi_0\sim\mathcal{N}(m', \sigma')$
\State $\xi(t_0) \gets \xi_0$
\LineComment sample the total variation
\State $K\sim(\mathcal{N}(m'', \sigma''))^2$
\State $y_1,...,y_{N-1}\sim\mathbb{U}([0, K])$
\State $y_0\gets0$, $y_n\gets K$
\State orderAndRename($y_0, \ldots, y_n$) 
\LineComment now $y_1 \leq y_2 \leq...\leq y_{N-1}$
\State $s_0\sim\text{Discr}(-1, 1)$
\While{$i\leq N$}
\State $s\gets$Binomial($q$) \Comment $P(s=-1) = q$
\State $s_{i+1} = s_i\cdot s$ 
\State $\xi(t_{i+1}) = \xi(t_i) + s_{i+1}(y_{i+1}-y_i)$ 
\EndWhile
\end{algorithmic}
\end{algorithm}
\vspace{-1.35cm}
\end{minipage}
\end{wrapfigure}
Intuitively, $\mu_0$ makes \emph{simple} trajectories more probable, considering total variation and number of changes in monotonicity as indicators of complexity of signals. The measure $\mu_0$, operating on piece-wise linear functions over the interval $\mathcal{I} = [a, b]$ (which is a dense subset of the set of continuous functions over $\mathcal{I}$), can be algorithmically defined as in Algorithm~\ref{alg:mu0} (default parameters are set as $a=0, b=100, \Delta=1, m'=m''=0'', \sigma'=\sigma''=1, q=0.1$).
Note that, although the feature space $\mathbb{R}^{\mathcal{T}}$ (which we call the \emph{latent semantic space}) into which $\rho$ (and thus Equation (\ref{eq:stl-kernel})) maps formulae is infinite-dimensional, in practice the kernel trick allows to circumvent this issue by mapping each formula to a vector of dimension equal to the number of formulae which are in the training set used to evaluate the kernel (Gram) matrix.


\paragraph{Bayesian Optimization (BO)} is a widely-used framework for black-box optimization problems, e.g. of the form $\bm{x}^{\star}=\text{argmax}_{\bm{x}\in\mathcal{X}} f(\bm{x})$ being $\mathcal{X}\subset \mathbb{R}^d$, which consists in sequentially at each iteration $t$ observing noisy evaluations of the objective function $y_t=f(\bm{x}_t) + \varepsilon_t$ where the noise is $\varepsilon_t\sim \mathcal{N}(0, \sigma^2)$, at candidate points $\bm{x}_t$ chosen by an acquisition function. 
The Gaussian Process Upper Confidence Bound (GP-UCB) algorithm consists in interpolating the available observations using a Gaussian Process (GP, i.e. a Bayesian non-parametric regression approaches) \cite{gp}, which is the so-called emulation phase, and adding candidate points to the training set of the GP in an iterative fashion using Upper-Confidence Bound (UCB, \cite{gp-ucb}) as acquisition function. More specifically it considers $f$ to be a sample path from a GP with $0$ mean and stationary kernel function $k: \mathcal{X}\times \mathcal{X}\mapsto \mathbb{R}$, i.e. $f\sim GP(0, k)$, so that at each iteration the posterior mean and variance of $f(\bm{x})$ are derived, and respectively denoted as $\mu_{t-1}(\bm{x})$ and $\sigma^2_{t-1}(\bm{x})$. The UCB acquisition function selects the next candidate point to be evaluated as $\bm{x}_t = \text{argmax}_{x\in \mathcal{X}} \mu_{t-1}(\bm{x}) + \beta_t^{\frac{1}{2}}\sigma_{t-1}(\bm{x})$, with $\beta_t\in \mathbb{R}$.

\section{Retrieval-Augmented STL Requirement Mining}\label{sec:method}

The problem we address with our proposed methodology is that of mining a STL specification from a dataset of labeled trajectories  (where labels indicate whether the system complies to some desired behaviour), without the need of inferring or knowing a generative model of the system. This is a supervised two-class classification problem, in which the input is a set of trajectories partitioned in the subset of those labeled as regular (or positive) and those which are anomalous (or negative), denoted respectively as $\mathcal{D}_p$ and $\mathcal{D}_n$, and the output is a (set of) STL formula(e) able to discriminate among the two sets. We also assume that such datasets come from unknown stochastic processes $\bm{X}_p$ and $\bm{X}_n$, respectively. 
We adopt the approach of \cite{roge} and optimize the following function in order to mine a STL formula $\varphi$ with the above desiderata:
\begin{equation}
    G(\varphi) = \frac{\mathbb{E}[R_{\varphi}|\bm{X}_p] - \mathbb{E}[R_{\varphi}|\bm{X}_n]}{\sigma(R_{\varphi}|\bm{X}_p) + \sigma(R_{\varphi}|\bm{X}_n)}
    \label{eq:optim}
\end{equation}
with $R_{\varphi}$ defined in Equation~\ref{eq:stochrob}. The intuition behind Equation~\ref{eq:optim} is that the higher the value of $G(\varphi)$ (i.e. the bigger the difference between the average robustness of the two stochastic processes, relative to the length scale given by their standard deviation $\sigma$), the more robustly the inferred $\varphi$ will separate the sets of positive and negative trajectories. 
The key insights of our approach are that: (i) we frame the learning problem as the optimization of Equation~\ref{eq:optim} in the latent semantic space of formulae, i.e. in the space of embeddings of formulae individuated by the STL kernel of Equation~\ref{eq:stl-kernel}; (ii) we construct a dense (hierarchical) vector database containing the embeddings of millions of STL formulae, that we query anytime we need to invert the embeddings resulting from the previous step. This IR step approximates the inverse of Equation \ref{eq:stl-kernel} by finding specification whose latent representation is the nearest to the searched embedding (among the ones in the database), to obtain STL formulae for evaluating $G(\varphi)$. Hence the focus of our procedure is on inferring the semantic features a formula needs to have to discriminate between positive and negative trajectories, without a-priori constraints on its syntactic structure.  

Summarizing, the proposed methodology consists in interleaving steps of Bayesian Optimization (BO) and Information Retrieval (IR): the details of these two building blocks are given in Sections~\ref{subsec:bo} and \ref{subsec:db}, respectively.

\subsection{Building and Querying the Semantic Vector Database}\label{subsec:db}
The STL kernel of Equation~\ref{eq:stl-kernel} allows to map formulae from the discrete space of their syntactic representation to a continuous space preserving their semantics (i.e. in the individuated latent space semantically similar formulae are mapped to nearby representations), however such map is not invertible. Given the recent advances in indexing structures and search mechanisms for dense vector databases \cite{ir-survey}, often accompanied by open-source software, we devise a simple yet effective strategy for inverting the STL kernel representations: storing in a reasoned and hierarchical database the embeddings of millions of formulae (more precisely of pairs containing both the formula and the corresponding representation), that we call \textit{semantic vector database}, and using Approximate Nearest Neighbors (ANN) search techniques to obtain STL requirements starting from a candidate embedding.

\paragraph{Selecting STL Formulae} In order to build such database we first need to select the set of formulae to store: we do so by fixing the maximum number $M$ of nodes and the maximum number of variables $N$ (i.e. the maximum dimensionality of the signals) allowed, then we enumerate all templates (i.e. STL formulae where constants are replaced by parameters) satisfying those constraints as detailed in Algorithm~\ref{alg:stl-templates}.
\begin{algorithm}[t]
\begin{adjustbox}{minipage=\linewidth,scale=0.95}
\caption{Algorithms for generating STL formulae templates}
\label{alg:stl-templates}
\begin{algorithmic}
\Require $M$, $N$
\Ensure all\_phis \Comment{\text{STL templates of formulae with max $N$ vars and $M$ nodes}}
\State $\text{all\_phis}\gets []$
\State all\_phis.append(generateAtomicPropositions())  \Comment{$x_i\leq 0$ or $x_i\geq 0$, $\forall i\leq N$}
\For{$2\leq m\leq M$}
\State $\text{prev\_phis}\gets$getPhisGivenNodes($m-1$) \Comment{retrieve templates with $m-1$ nodes}
\State $\text{unary\_ops}\gets$expandByUnaryOperators(prev\_phis) \Comment{$F$, $G$, $\neg$}
\State all\_phis.append(unary\_ops)
\State l\_list, r\_list $\gets$getPairsGivenSum($m$) \Comment{all pairs $(l, r): l+r=m, l\leq r$}
\For{$(l, r)\in$ [l\_list, r\_list]}
\State l\_phis$\gets$getPhisGivenNodes($l$)
\State r\_phis$\gets$getPhisGivenNodes($r$)
\State binary\_ops$\gets$expandByBinaryOpeators(l\_phis, r\_phis) \Comment{$\wedge$, $\vee$, $U$}
\State all\_phis.append(binary\_ops)
\EndFor
\EndFor
\end{algorithmic}
\end{adjustbox}
\end{algorithm}
Once all the templates have been generated, we instantiate each of them on a set of parameter configurations $\mathcal{P} = \{p_0,\ldots p_{|\mathcal{P}|}\}$ to be filtered by Signature-based Optimization as defined in \cite{enumerate-stl}. Denoting as $\varphi(p)$ the application of the parameter set $p$ to the template formula $\varphi$, and given a set of trajectories $\hat{\mathcal{T}} = \{\xi_0,\ldots, \xi_{|\hat{\mathcal{T}}|}\}$ (with $p\in \mathcal{P}$, and $\hat{\mathcal{T}}\subset \mathcal{T}$), it consists in building a matrix $S\in \mathbb{R}^{|\mathcal{P}|\times |\hat{\mathcal{T}}|}$, whose rows are called the \textit{signature} of the formula, s.t. $S(i, j) = \rho(\varphi(p_i), \xi_j)$, i.e. the cell $(i, j)$ of such matrix holds the robustness of the $i$-th instantiation of $\varphi$ on the $j$-th trajectory. In such matrix, the distance between rows can be considered as a proxy for semantic similarity among different parametrizations of the same template, hence a selection on formulae can be done by keeping only those whose signature has a distance higher than a certain threshold $\tau\in\mathbb{R}_{\geq 0}$ from formulae already selected. This helps in removing redundancy in the final database structure.

\paragraph{Computing Formulae Embeddings} Once the list of formulae to be stored in the database is provided, the STL kernel embeddings of such specifications needs to be computed, as for Equation \ref{eq:stl-kernel}. In order to do so, one needs to (i) fix a probability measure on the space of trajectories and (ii) sample a set of $n_{\text{train}}$ formulae against which the kernel is computed. For the first point, we use the default measure $\mu_0$, computed as in Algorithm~\ref{alg:mu0}, proved to be quite general in \cite{stl-kernel}. For the second point we instead sample from a distribution $\mathcal{F}$ over STL formulae which is algorithmically defined by a syntax-tree random recursive growing scheme, that recursively generates the nodes of a formula given the probability $p_{\mathit{leaf}}$ of each node being an atomic predicate, and a uniform distribution over the other operator nodes. We remark that the $n_{\text{train}}$ determines, by construction, the dimension of the vectors stored in the database. 

\paragraph{Index Structure Organization} It is worth noting that the semantic vector database is built once (although it can be further extended) and leveraged in all the applications, hence it is fundamental to organize it in a reasoned way, given the different contexts it can be deployed in. In particular, we hierarchically index the embeddings based on the number of different variables and the number of nodes appearing in the formula. The rationale behind the first choice is that not all signals that we wish to classify with our methodology have the same dimension, the second choice moves instead from the point of  view of interpretability: the smaller the syntax tree of a formula, the higher its human-understandability, hence we incorporate in the retrieval strategy the preference for syntactically simpler formulae. 

\paragraph{Semantic Database Implementation} For indexing the embeddings, we use the tools provided by the FAISS Python library \cite{faiss-gpu}, which allows us to leverage GPU acceleration for both indexing and querying steps. Since the amount of vectors we need to store is of the order of millions, we opt for the Inverted File Index with Product Quantization (IVFPQ) to accelerate the search procedure (numerical results will follow in the remainder of this Section), which is a composite index that both partitions the embeddings in Voronoi cells and reduces their dimension using Product Quantization \cite{ir-survey}. Euclidean (or $L_2$) distance is used as similarity measure between vectors.
We remark that the framework we propose is flexible w.r.t. the choice of the index structure, hence this component can be replaced with any similarity search engine on vector databases without altering the overall structure of the algorithm.

\begin{table}[t]
\caption{Quantiles of $AP@K$, $NDCG@K$ and kernel similarity $k(\varphi, \hat{\varphi})$ between the searched formula $\varphi$ and the top result $\hat{\varphi}$, across $1000$ queries to the semantic database, varying the number of nodes in the syntactic tree of the searched formulae.}\label{tab:ir-results}
\centering
\resizebox{1.\linewidth}{!}{
\begin{tabular}{l|llll|llll|llll}
\toprule
 \multicolumn{5}{c}{$AP@K$} & \multicolumn{4}{c}{$NDCG@K$} & \multicolumn{4}{c}{$k(\varphi, \hat{\varphi})$}\\
\midrule
$n_{\text{nodes}}$ &  1quart &   median &   3quart &  99perc &  1quart &   median &   3quart &  99perc & 1quart &   median &   3quart &  99perc \\
$(0, 5]$ & $0.804$ & $1.000$ & $1.000$ & $1.000$ & $1.000$ & $1.000$ & $1.000$ & $1.000$ & $0.810$ & $0.901$ & $0.976$ &  $0.992$ \\
$(5, 10]$  & $0.763$ & $1.000$ & $1.000$ & $1.000$ & $1.000$ & $1.000$ & $1.000$ & $1.000$ & $0.774$ & $0.883$ & $0.946$ & $0.979$ \\
$(10, 15]$  & $0.684$ & $0.932$ & $1.000$ & $1.000$ & $1.000$ & $1.000$ & $1.000$ & $1.000$ & $0.728$ & $0.854$ & $0.928$ & $0.972$ \\
$(15, 20]$  & $0.635$ & $0.873$ & $1.000$ & $1.000$ & $1.000$ & $1.000$ & $1.000$ & $1.000$ & $0.742$ & $0.848$ & $0.917$ & $0.976$ \\
\bottomrule
\end{tabular}
}
\end{table}
In order to build the index we enumerate all formulae templates following Algorithm~\ref{alg:stl-templates} with $M=5$ and $N=3$. Given that the measure $\mu_0$ is used for computing the embeddings, we keep into account the typical range of its trajectories (see Figure~\ref{fig:mu0}) for setting the variable thresholds, hence using the grid of $10$ equally-spaced points between $-4$ and $4$ as possible parameter instances; for the temporal operators thresholds, we instead leverage the fact that signals for computing the STL kernel representations have $100$ points and use the grid of $10$ equally-spaced points between $0$ and $100$ as possible instances for both left and right time-bounds. We select the formulae to store in the index by signature-based optimization with cosine distance and threshold $\tau = 0.9$. Finally, as done in \cite{stl-kernel}, we use $n_{\text{train}}=1000$ STL formulae for computing the embeddings. This leads to a total of $\sim \num{10000000}$ stored formulae, that we organize in a total of $6$ indexes: first splitting them based on the number of variables they contain (namely $1$, $2$ or $3$), then splitting each of them based on the maximum number of nodes allowed (namely $4$ or $5$). Of course, alongside the embeddings, we store the STL formulae organized in a list structure mirroring the organization of the corresponding representations. Searching for $1000$ formulae with $3$ variables on such index (hence scanning all the database) requires $1\pm 0.5s$, while computing their embeddings takes $6\pm 2s$ exploiting GPU acceleration, finally the whole process of creating the database takes $\sim 20h$ on a AMD EPYC $7542$ machine with  $64$ GB of RAM and a NVidia A$100$ with $80$GB of memory. Moreover, storing the index requires $\sim 180MB$, which is order of magnitudes less then the $\sim 70GB$ required in case plain embeddings should be stored (e.g. for exact search). 

\paragraph{Semantic Database Effectiveness} In order to assess the effectiveness of our information retrieval system, we use Average Precision at $K$ ($AP@K$) and Normalized Discounted Cumulative Gain at $K$ ($NDCG@K$), to measure both the relevancy of the results and the ranking quality. 
Defining the precision at $K$ (denoted as $P@K$) as the number of relevant results among the top $K$ retrieved, and $rel_k$ as the indicator function being $1$ when the $k$-th ranked result is relevant and $0$ otherwise, then $AP@K=\frac{\sum_{k=1}^K (P@K\cdot rel_k)}{\sum_{k=1}^K rel_k}$. If we instead define $r_k$ as the \textit{graded relevance} (or gain) of the result at position $k$, i.e. an integer score indicating its relevance,
then the Discounted Cumulative Gain at K is $DCG@K=\frac{\sum_{k=1}^K r_k}{\log_2(k+1)}$; the Ideal $DCG@K$ ($IDCG@K$) is the maximum achievable $DCG@K$ given a set of relevance scores and finally $NDCG@K=\frac{DCG@K}{IDCG@K}$. 
The definitions of $AP@K$ and $NDCG@K$ given above are based on the concept of \textit{relevancy} of the result, which is not immediate to state in our scenario. Intuitively, we consider a retrieved formula $\hat{\varphi}$ semantically similar to a searched formula $\varphi$ if they behave similarly across an arbitrary number of trajectory, i.e. if they are (un)satisfied on the same subset of signals. Hence, given a similarity threshold $\omega\in [0, 1]$ and a set of trajectories $\{\xi_j\}_{j=1}^{n_{\text{traj}}}\subset \mathcal{T}$, we say that $\hat{\varphi}$ is a relevant result w.r.t. the query $\varphi$ if $\frac{|\{j: s(\varphi, \xi_j) = s(\hat{\varphi}, \xi_j)\}|}{n_{\text{traj}}}\geq \omega$. Moreover, we also consider the kernel similarity of Equation~\ref{eq:stl-kernel} (whose maximum is $1$) as an indicator of relevancy of the results, which accounts also for the quantitative semantics of formulae. 
In our experiments, we used $n_{\text{traj}}=\num{10000}$ signals sampled from $\mu_0$, $\omega=0.9$ and STL formulae sampled from $\mathcal{F}$ with $N=3$, varying the number $n_{\text{nodes}}$ of nodes allowed in the syntax tree of the formula. Results in terms of $AP@K$ and $NDCG@K$ (with $K=5$) are reported in Table~\ref{tab:ir-results}, highlighting both the ranking quality of our methodology ($NDCG@K$ always $1$) and the relevancy of retrieved results (median $AP@K > 0.87$ in the worst case), as well as resilience to the size of the searched formulae (we recall that the database contains specification with $n_{\text{nodes}}\leq 5$). Moreover, we also observe that the kernel similarity between the query $\varphi$ and the top retrieved formula $\hat{\varphi}$ is high (median above $0.84$ in all tested cases) across all specification dimensions.

\subsection{Bayesian Optimization in the Semantic Space of Formulae}\label{subsec:bo}

\begin{algorithm}[t]
\begin{adjustbox}{minipage=\linewidth,scale=0.95}
\caption{STL Requirement Mining Combining BO and IR}
\label{alg:mining}
\begin{algorithmic}
\Require $\mathcal{D}_p$, $\mathcal{D}_n$, SemanticDB  
\Ensure $\tilde{\varphi}$ 
\State k\_phis, phis, g\_phis $\gets getInitialCandidates(\mathcal{D}_p, \mathcal{D}_n)$
\State $i\gets 0$
\State opt$\gets [0, g\_phis]$
\While{ $i\leq maxiter$ or $(opt[i] - opt[i-1]) < \varepsilon$}
\State new\_phis$\gets$optimizeUCB()
\State k\_phis.append(new\_phis)
\State phis.append(retrieveFromEmbedding(new\_phis, SemanticDB))
\State g\_phis.append(evaluteObjective(new\_phis, $\mathcal{D}_p$, $\mathcal{D}_n$))
\State opt.append($\max(g\_phis)$)
\EndWhile 
\State $\tilde{\varphi}\gets$ phis[\text{argmax}(g\_phis)]
\end{algorithmic}
\end{adjustbox}
\end{algorithm}
As already mentioned, the goal of the methodology we propose is to learn the function mapping semantic embeddings of STL formulae (built as for Equation~\ref{eq:stl-kernel}) to values of $G(\varphi)$ of Equation~\ref{eq:optim}, given labeled datasets $\mathcal{D}_p$ and $\mathcal{D}_n$ of respectively positive and negative trajectories. This implies searching in the latent semantic space for the continuous representation of a formula able to discriminate between signals belonging to $\mathcal{D}_p$ and $\mathcal{D}_n$. Being it a non-linear non-convex optimization problem, we tackle it by means of Bayesian Optimization (BO), and more specifically of the Gaussian Process Upper-Confidence Bound (GP-UCB) algorithm~\cite{gp-ucb}. We recall that the typical BO loop consists in: (i) fit the GP on the available data (pairs $(k(\varphi), G(\varphi))$ in our case, with $\varphi$ STL formula and $k(\varphi)$ its STL kernel embedding); (ii) optimize the UCB acquisition function to find the best candidate (the kernel embedding $k(\hat{\varphi})$ of a formula $\hat{\varphi}$ improving Equation~\ref{eq:optim}); (iii) query the objective function, i.e. Equation~\ref{eq:optim}, on the new points to add new data pairs to the training set. In order to obtain an STL formula $\hat{\varphi}$ starting from its emebedding $k(\hat{\varphi})$ (i.e. between the steps (ii) and (iii) of the BO loop), we augment the procedure with a retrieval step, querying the semantic vector database on $k(\hat{\varphi})$. An overview of the overall methodology is reported in Algorithm~\ref{alg:mining}. In our experiments we use GP with Matern kernel (starting with an initial batch of only $10$ points) and optimize the UCB acquisition function using Stochastic Gradient Descent (SGD), given the high-dimensionality of the latent semantic space (namely $\num{1000}$, as detailed in Section \ref{subsec:db}); this is done in Python exploiting the BoTorch library \cite{botorch}, which allows to leverage GPU acceleration.

\section{Experiments}\label{sec:exp}
\subsection{Experimental Setting}\label{subsec:setting}
We test our proposed algorithm across a number of benchmark datasets found in the related literature (see Section \ref{sec:related}). Unless differently specified, we perform $5$-fold cross validation and report mean and standard deviation of the results across the folds. The metrics we use for assessing the quality of the results are: Misclassification Rate ($MCR$), Precision ($Prec$) and Recall ($Rec$), defined as follows, given positive and negative trajectory datasets $\mathcal{D}_p$ and $\mathcal{D}_n$ and the mined STL specification $\hat{\varphi}$:
\begin{small}
\begin{align*}
    MCR &=& \frac{|\{\xi_j\in\mathcal{D}_p : s(\hat{\varphi}, \xi_j) < 0\}| + |\{\xi_j\in\mathcal{D}_n : s(\hat{\varphi}, \xi_j) > 0\}|}{|\mathcal{D}_p| + |\mathcal{D}_n|} \\
    Prec &=& \frac{|\{\xi_j\in\mathcal{D}_p : s(\hat{\varphi}, \xi_j) > 0\}|}{|\{\xi_j\in\mathcal{D}_p : s(\hat{\varphi}, \xi_j) > 0\}| + |\{\xi_j\in\mathcal{D}_n : s(\hat{\varphi}, \xi_j) > 0\}|} \\
    Rec &=& \frac{|\{\xi_j\in\mathcal{D}_p : s(\hat{\varphi}, \xi_j) > 0\}|}{|\{\xi_j\in\mathcal{D}_p : s(\hat{\varphi}, \xi_j) > 0\}| + |\{\xi_j\in\mathcal{D}_p : s(\hat{\varphi}, \xi_j) < 0\}|}
\end{align*}    
\end{small}
In the above Equation, using the jargon of binary classification algorithms, the true positives are $TP=\{\xi_j\in\mathcal{D}_p : s(\hat{\varphi}, \xi_j) > 0\}$, the true negatives are $TN=\{\xi_j\in\mathcal{D}_n : s(\hat{\varphi}, \xi_j) < 0\}$, false positives are instead defined as $FP=\{\xi_j\in\mathcal{D}_n : s(\hat{\varphi}, \xi_j) > 0\}$ and finally false negatives $FN=\{\xi_j\in\mathcal{D}_p : s(\hat{\varphi}, \xi_j) < 0\}$.

The retrieval step of Algorithm~\ref{alg:mining} requires to normalize trajectories of $\mathcal{D}_p$ and $\mathcal{D}_n$ by removing the mean and diving by the standard deviation of each dimension, in order to align their domain with that of $\mu_0$ (i.e. the measure w.r.t. which semantic embeddings are computed). At the end of the procedure, they are scaled back to their original range, and the same inverse normalization is applied to the thresholds of the variables appearing in the mined specification (since all the formulae in the semantic vector database have thresholds compatible with $\mu_0$ by construction). Another aspect to take into account when dealing with the retrieval step of Algorithm~\ref{alg:mining} is that of time parameters: the sampling rate of $\mu_0$ is $1$ per time-step, considering a total of $100$ points, but this might vary across trajectory distribution. Assuming that the target distribution $\mathcal{D}_{\text{test}}$ has $n_{\text{test}}$ time points, without loss of generality we can consider them to be sampled at rate of $1$ per time step, so that a time interval $[a, b]$ referred to a trajectory sampled from $\mu_0$ can be mapped to the corresponding time interval $[a_{\text{test}}, b_{\text{test}}]$ for a trajectory sampled from $\mathcal{D}_{\text{test}}$ as: $[a_{\text{test}}, b_{\text{test}}] = [\left \lfloor a\cdot (n_{\text{test}}/100) \right \rfloor, \left \lfloor a\cdot (n_{\text{test}}/100) \right \rfloor + \left \lceil (b-a)\cdot (n_{\text{test}}/100) \right \rceil]$. Moreover, before querying the semantic vector database, we restrict the search space to only those indexes containing formulae having a number of variables compatible with the dimension of $\mathcal{D}_{\text{test}}$ and, for the sake of interpretability, we first search on indexes containing specifications with at most $4$ nodes in the syntactic tree and, if the outcome of the procedure is not satisfactory, we allow also to search for more complex requirements (as we will report in the remainder of this Section, this has not been necessary in any of test cases). All the experiments are implemented in Python, leveraging GPU acceleration in every sub-procedure thanks to the PyTorch~\cite{pytorch} library, and tested on a AMD EPYC $7542$ machine with  $64$ GB of RAM and a NVidia A$100$ with $80$GB of memory. 
\subsection{Case Studies}\label{subset:results}
\begin{figure}[t]
    \centering
    \begin{adjustbox}{minipage=\linewidth,scale=0.8}
    \subfigure[Linear System]{
        \includegraphics[width=0.3\textwidth]{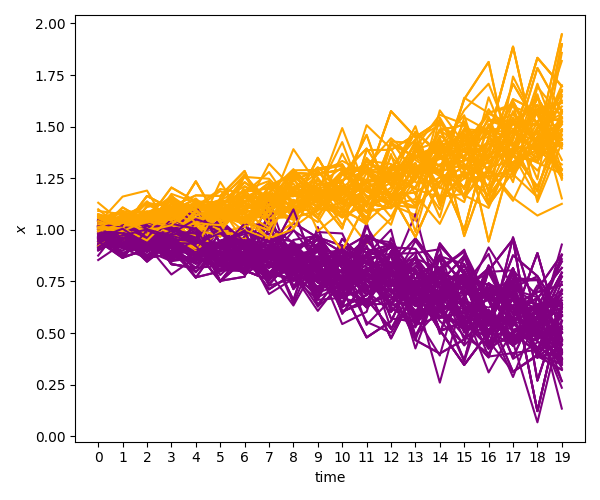}
        \label{subfig:linear}}
    \hspace{\fill}
    \subfigure[HAR]{
        \includegraphics[width=0.3\textwidth]{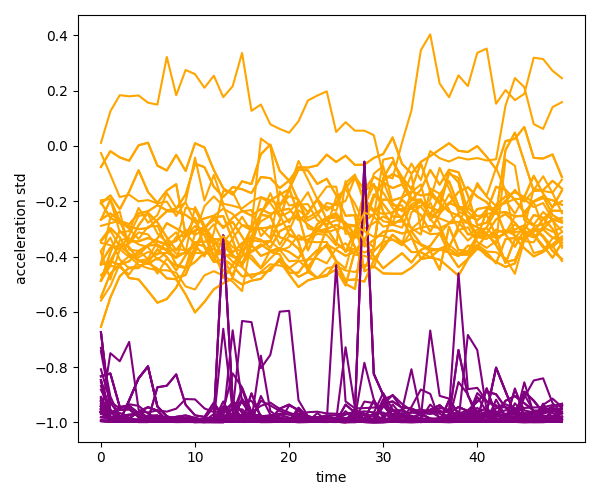}
        \label{subfig:har}}
    \hspace{\fill}
    \subfigure[Control of a train]{
        \includegraphics[width=0.3\textwidth]{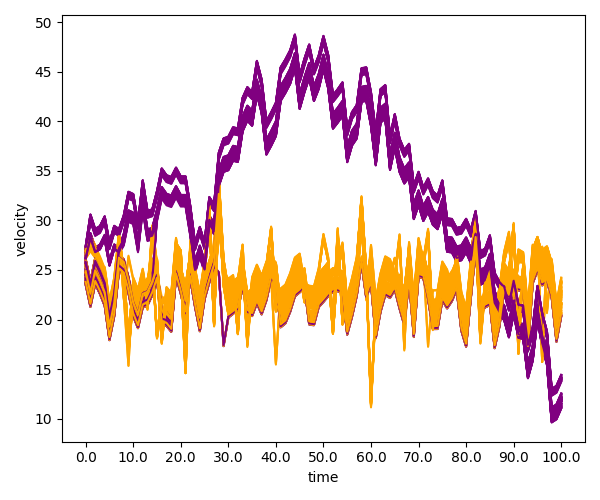}
        \label{subfig:train}}
    \hspace{\fill}
    \subfigure[Maritime]{
        \includegraphics[width=0.3\textwidth]{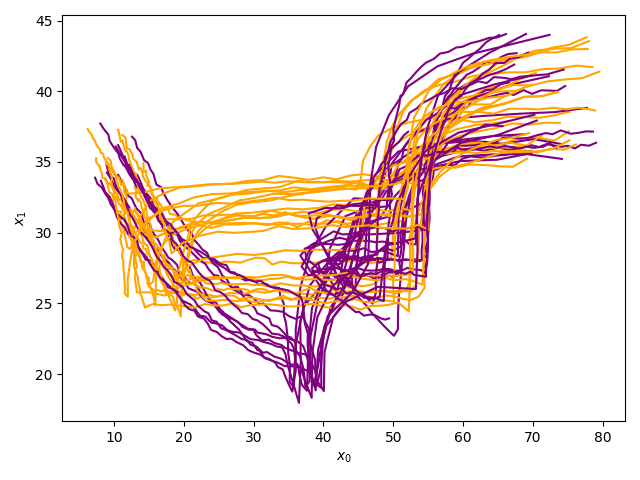}
        \label{subfig:naval}}
    \hspace{\fill}
    \subfigure[LP$5$]{
        \includegraphics[width=0.6\textwidth]{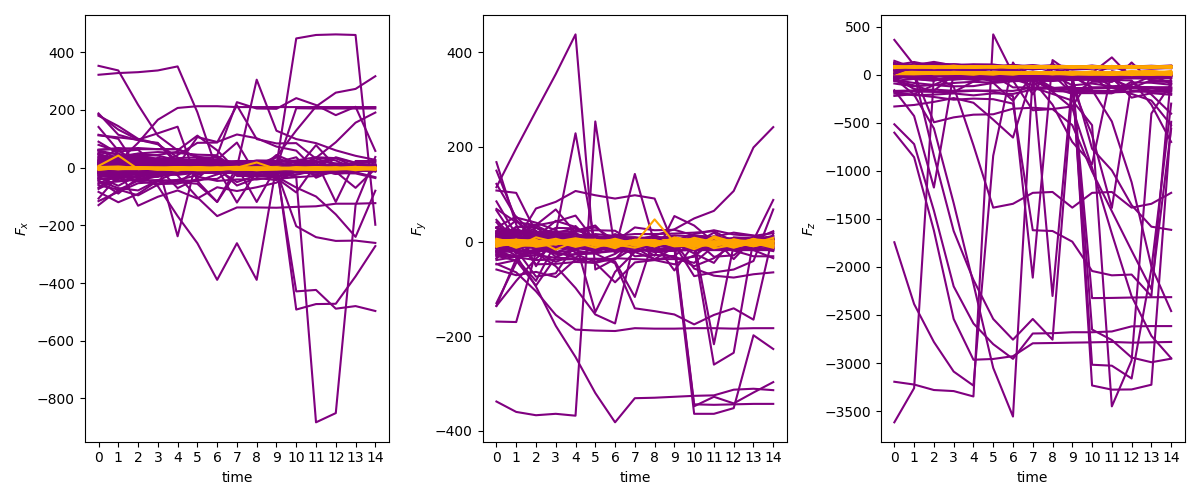}
        \label{subfig:robot}}
    \end{adjustbox}
    \caption{\textcolor{orange}{Regular} and \textcolor{violet}{anomalous} trajectories sampled from the testes datasets.}
    \label{fig:traj-1-var}
\end{figure}

\paragraph{Linear System} In \cite{enumerate-stl} the authors generate trajectories using the following dynamical system: regular trajectories evolve as $\dot{x}=0.03\cdot x + w$ while anomalous ones as $\dot{x} = -0.03\cdot x + w$, being $w(t)$ a white noise with variance $0.04$. Following their approach, we generated $100$ positive and $100$ negative examples, sampling $100$ points for each trajectory, a random subset of the obtained signals is reported in Figure~\ref{subfig:linear}.
\paragraph{Maritime surveillance} The \textit{Maritime Surveillance} dataset is a common benchmark for STL anomaly detection, used e.g. in \cite{decision-tree,enumerate-stl,roge}. It consists in $2$-dimensional signals reporting vessels behaviour, divided in $1000$ regular and $1000$ anomalous examples, each consisting of $61$ sampling points. Random signals belonging to the maritime dataset are reported in Figure~\ref{subfig:naval}.
\paragraph{Human Activity Recognition (HAR)} The HAR dataset created in \cite{human-data} and publicly available at \cite{ml-repo} contains recording of humans performing activities in daily livings. Following the approach \cite{enumerate-preprint}, we split the $6$ possible activities in $3$ static postures and $3$ dynamic activities, and use these as classification labels (for a total of $36$ and $58$ trajectories belonging to the mentioned classes, respectively). Signals belonging to this dataset are $1$-dimensional, representing the standard deviation of the acceleration of the subjects in the $x$-direction at $50$ sampling points. A subset of positive and negative trajectories belonging to the HAR dataset is reported in Figure~\ref{subfig:har}. 
\paragraph{Robot Execution Failures in Motion with Part (LP$5$)} The LP$5$ dataset, available at \cite{ml-repo}, consists of $3$-dimensional signals, representing force measurements along the $3$ axes, each recorded at $15$ time steps. As in \cite{enumerate-preprint}, we consider as positive the $44$ instances showing normal behaviour and as negative the $26$ exhibiting bottom collisions. Trajectories belonging to this dataset are shown in Figure~\ref{subfig:robot}.
\paragraph{Cruise Control of Train} Following \cite{enumerate-stl}, we test our procedure on the dataset coming from the (noisy) observations of the velocity of a train, simulated in a cruise control scenario. Such signals are $1$-dimensional and sampled for $100$ points; classes contain $100$ trajectories each. A random sample of such trajectories is shown in Figure~\ref{subfig:train}.

\begin{table}[t]
\caption{Best mined formula, mean and std of MCR, Prec and Rec in the test set across the folds for each case study, computational time (in seconds) required to find the reported formula using Algorithm \ref{alg:mining}.}\label{tab:mining-results}
\centering
\begin{tabular}{l|lllll}
\toprule
 \multicolumn{2}{c}{$\hat{\varphi}$} & $MCR$ & $Prec$ & $Rec$ & Time ($s$)\\
\midrule
Linear & $F_{[70, \infty]} (x_0 \leq 1.16)$ & $0.00\pm 0.00$ & $1.00\pm 0.00$ & $1.00\pm 1.00$ &  $7.32\pm 1.45$\\
Maritime & $(x_1 \geq 23.19) U (x_0\leq 32.56)$ & $0.07\pm 0.01$ & $0.97\pm 0.01$ & $1.00\pm 0.00$ & $12.4\pm 2.63$\\
HAR  & $G (x_0\geq -0.75)$ & $0.00\pm 0.00$ & $1.00\pm 0.00$ & $1.00\pm 0.00$ & $8.32\pm 0.25$ \\
LP$5$  & $G_{[4, \infty]} (x_2\geq 0.084)$ & $0.01\pm 0.01$ & $0.98\pm 0.01$ & $1.00\pm 0.00$ & $11.6\pm 1.44$ \\
Train  & $G_{[0, 36]} (x_0\leq 37)$ & $0.03\pm 0.01$ & $0.93\pm 0.02$ & $1.00\pm 0.00$ & $8.21\pm 1.23$\\
\bottomrule
\end{tabular}
\end{table}

\subsubsection{Discussion}
\begin{wrapfigure}{rh!}{0.38\textwidth}
\vspace*{-0.5cm}
\includegraphics[scale=0.35]{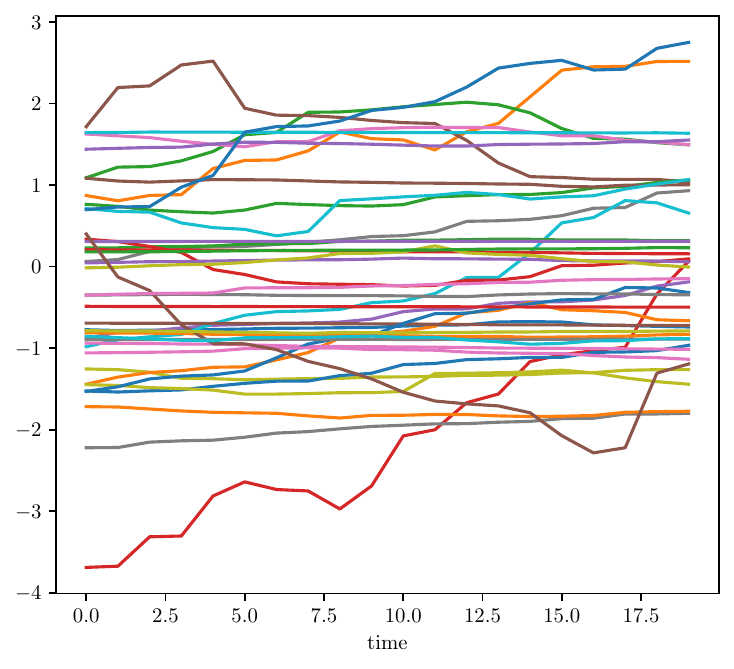}
\vspace*{-0.5cm}
\caption{Signals sampled from $\mu_0$.}
\label{fig:mu0}
\vspace*{-0.7cm}
\end{wrapfigure}
Results of the retrieval-augmented STL requirement mining procedure are shown in Table~\ref{tab:mining-results}, where we report the best formula we infer (denoted as $\hat{\varphi}$), along with performance metrics and computational time for each case study. In this context, with \textit{best} formula we denote the mined formula having the lowest $MCR$ in the test-set, using the lower number of nodes as tiebreaker.
Notably, we can observe that all mined requirements are interpretable, having a syntax tree of at most $3$ nodes, in line with more recent related works (namely \cite{enumerate-preprint,enumerate-stl,roge}), and act as almost-perfect discriminators among regular and anomalous trajectories (see $MCR$ in Table~\ref{tab:mining-results}). Moreover, we can observe that the mined specifications never produce false positives (reporting a recall of $1$ in every test case). It is worth noting that even if the LP$5$ benchmark is $3$-dimensional, the mining procedure outputs a formula in which only a variable appears (in line with \cite{enumerate-preprint}). 
We find it interesting to remark the difference between trajectories sampled from the measure $\mu_0$ used to compute STL embeddings (a random sample of signals belonging to this distribution is shown in Figure~\ref{fig:mu0}) and those of the test benchmarks (reported in Figure~\ref{fig:traj-1-var}), to underline the applicability of our methodology across various trajectories distribution, without the need of recomputing the semantic vector database on each data distribution. 
Finally, another relevant difference between our approach and related works is that we fully exploit GPU acceleration in every step of the mining procedure, resulting in lower computational times. Indeed as reported in Table~\ref{tab:mining-results}, our method is able to learn a classifier in less than $15s$ for all the tested cases, while \cite{enumerate-stl} reports a computational time of $39.05s, 23.17s, 32.32s$ for the Linear, Maritime and Train datasets, respectively; \cite{enumerate-preprint} register $22.07s$ and $22.83s$ for mining a discriminating formula in the HAR and LP$5$ banchmarks; finally \cite{roge} takes $73s$ to classify signals belonging to the Maritime dataset.

\section{Related Work}\label{sec:related}
\paragraph{Mining STL specifications from data} has seen a surge of interest in the last few years, as reported in \cite{stl-mining-survey}. At a high level, following \cite{stl-mining-survey}, contributions in this field can be characterized by: template-based vs template-free procedures (in the first, the mining consists in finding the optimal parameters for a given syntactic skeleton, while in the second both the structure and the parameters of the specification have to be inferred), supervised vs unsupervised approaches (following the availability of labeled trajectory datasets) and online vs offline methodology (the first concerns updating the inferred requirements every time new data are available, the latter considers instead a static set of data). In this scenario our algorithm can be defined as a template-free, supervised, offline approach (besides, it can be cast as an unsupervised learning problem by just changing the objective function of Equation \ref{eq:optim}, keeping intact the structure of Algorithm \ref{alg:mining}). Most of the related works with these characteristics decompose the task in two steps, i.e. as bi-level optimization problem: they first learn the structure of the specification from data, and then instantiate a concrete formula using parameter inference methods. For the first step, a popular extension of STL, called Parametric Signal Temporal Logic (PSTL) has been devised \cite{pstl}, in which parameters replace both threshold constants in numerical predicates and time bounds in temporal operators. In \cite{grid-based} STL specifications are mined using a grid-based discretization of both time and value domain, used to guide a clusterization of the input signals, for assigning to each cluster a descriptive STL formula and obtain the final requirement as disjunction of cluster formulae. In \cite{decision-tree} a map between a fragment of PSTL and binary decision trees is established, so that the requirement mining procedure is cast into a classification problem tackled by decision tree, and then translation of the tree to a formula is done to infer the final specification. In \cite{enumerate-stl} STL requirements are mined using systematic enumeration of PSTL templates (in increasing order w.r.t. the syntactic size) and accuracy-driven instantiation of each template (towards optimizing logical classification), guaranteeing that the inferred formula is the smallest one reaching a given accuracy threshold. Another line of works \cite{gp-two,roge,gp-one} use instead Genetic Algorithms (GA) to learn the PSTL structure of the specification, and find optimal parameters by Gaussian Process Upper Confidence Bound (GP-UCB); notably \cite{roge} do not put any restrictions on the fragment of PSTL used. Our work differentiate from the cited papers in that it learns simultaneously both the structure and the parameters of the STL requirement, by performing optimization in a continuous space representing the semantics of formulae. 
\paragraph{Data Mining for Time Series Data}
Typically, anomaly detection and classification in time series data is done by either autoregressive statistical models, distance and density-based clustering methods, or prediction and reconstruction-based deep neural networks \cite{anomaly-survey,class-survey}. Such methods often require to individuate a latent representation of trajectories or to hand-craft features for the analysis, leading to a black-box solution to the problem, which do not offer any insight on the system at hand. Notably, there are works focusing on building time series classifiers which are interpretable, e.g. \cite{interpret-ts} that uses Symbolic Fourier Approximation to infer a symbolic representation of the input before learning a linear classifier, and \cite{interpret-ts-2} that uses saliency maps to characterize classes. Using STL as formal language for the learned class discriminator allows to describe emergent properties of a system in a rich yet interpretable way, leading to knowledge discovery about the process under analysis \cite{stl-mining-survey}.

\paragraph{Information Retrieval and Data Mining,} as well as their interaction, witnessed a growth of interest recently \cite{dm-ir-overview}. In particular, the adoption of the so-called dense vector databases (i.e. specialized storage systems designed for efficient management of dense vectors and supporting advanced similarity search) and retrieval methods has been proven to be effective in injecting factual information into Machine Learning (ML) techniques \cite{retrieval-ml}. Using information retrieval (IR) systems to support ML models access abstractly-represented knowledge stores helps them in generalize beyond training data, without the need of increasing their size, and notably helps for the interpretability and explainability of the outcomes, being inference grounded on retrieved information, often stored in a human-readable format \cite{retrieval-ml}.

\section{Conclusion}\label{sec:conclusion}
In this work we propose a methodology for mining STL requirements from time-series data, in a purely data-driven setting. In particular, we focus on binary classification of trajectories, such as those coming from observations of Cyber-Physical Systems. The core idea of our algorithm lies in the interaction between Bayesian Optimization and Information Retrieval techniques, by searching from the discriminating STL property directly in a continuous space representing its robust semantics. Doing so, we learn simultaneously the structure and the parameters of the specification, in such a way that it is possible to control its maximum allowed size, enhancing the interpretability of the inferred formula, to foster knowledge discovery about the system. Results of experiments on several benchmarks show that we achieve state-of-art accuracy, drastically reducing computational time w.r.t. related works, thanks to the exploitation of GPU acceleration in all steps of our methodology. It is also worth mentioning that by changing the optimization objective, our framework can be deployed in an unsupervised classification scenario, as well as in a requirement mining context in which only positive examples are available, extensions that we plan to implement as future work.





\bibliographystyle{splncs04}
\bibliography{biblio}
\end{document}